\documentclass[sigconf]{acmart}

\usepackage{CJKutf8}
\usepackage{subfigure}
\usepackage{balance}
\usepackage{multirow}

\newcommand{\ignore}[1]{}
\usepackage{tikz}

\AtBeginDocument{%
  }

\copyrightyear{2023}
\acmYear{2023}
\setcopyright{acmlicensed}
\acmConference[CIKM '23] {Proceedings of the 32nd ACM International Conference on Information and Knowledge Management}{October 21--25, 2023}{Birmingham, United Kingdom.}
\acmBooktitle{Proceedings of the 32nd ACM International Conference on Information and Knowledge Management (CIKM '23), October 21--25, 2023, Birmingham, United Kingdom}
\acmPrice{15.00}
\acmISBN{979-8-4007-0124-5/23/10}
\acmDOI{10.1145/3583780.3615119}

\begin{document}

%%
%% The "title" command has an optional parameter,
%% allowing the author to define a "short title" to be used in page headers.
\title{FlaCGEC: A Chinese Grammatical Error Correction Dataset with Fine-grained Linguistic Annotation}

%%
%% The "author" command and its associated commands are used to define
%% the authors and their affiliations.
%% Of note is the shared affiliation of the first two authors, and the
%% "authornote" and "authornotemark" commands
%% used to denote shared contribution to the research.
\author{Hanyue Du}
\affiliation{%
  \institution{East China Normal University}
  \city{Shanghai}
  \country{China}}
\email{hydu@stu.ecnu.edu.cn}

\author{Yike Zhao}
\affiliation{%
  \institution{East China Normal University}
  \city{Shanghai}
  \country{China}}
\email{ykzhao@stu.ecnu.edu.cn}

\author{Qingyuan Tian}
\affiliation{%
  \institution{East China Normal University}
  \city{Shanghai}
  \country{China}}
\email{qytian@stu.ecnu.edu.cn}

\author{Jiani Wang}
\affiliation{%
 \institution{East China Normal University}
 \city{Shanghai}
 \country{China}}
\email{jiani.wang@stu.ecnu.edu.cn}

\author{Lei Wang}
\affiliation{%
  \institution{Singapore Management University}
  \country{Singapore}}
\email{lei.wang.2019@phdcs.smu.edu.sg }

\author{Yunshi Lan}
\authornote{Corresponding author.}
\affiliation{%
  \institution{East China Normal University}
  \city{Shanghai}
  \country{China}}
\email{yslan@dase.ecnu.edu.cn}

\author{Xuesong Lu}
\affiliation{%
  \institution{East China Normal University}
  \city{Shanghai}
  \country{China}}
\email{xslu@dase.ecnu.edu.cn}

%%
%% By default, the full list of authors will be used in the page
%% headers. Often, this list is too long, and will overlap
%% other information printed in the page headers. This command allows
%% the author to define a more concise list
%% of authors' names for this purpose.
\renewcommand{\shortauthors}{Hanyue Du et al.}

\newcommand{\yscomment}[1]{\textcolor{orange}{[YS: #1]}}
\newcommand{\hycomment}[1]{\textcolor{blue}{[HY: #1]}}
\newcommand{\hyadd}[1]{\textcolor{blue}{#1}}

%%
%% The abstract is a short summary of the work to be presented in the
%% article.
\begin{abstract}
Chinese Grammatical Error Correction (CGEC) has been attracting growing attention from researchers recently.
In spite of the fact that multiple CGEC datasets have been developed to support the research, these datasets lack the ability to provide a deep linguistic topology of grammar errors, which is critical for interpreting and diagnosing CGEC approaches.
To address this limitation, we introduce FlaCGEC, which is a new CGEC dataset featured with fine-grained linguistic annotation.
Specifically, we collect raw corpus from the linguistic schema defined by Chinese language experts, conduct edits on sentences via rules, and refine generated samples manually, which results in $10$k sentences with $78$ instantiated grammar points and $3$ types of edits.
We evaluate various cutting-edge CGEC methods on the proposed FlaCGEC dataset and their unremarkable results indicate that this dataset is challenging in covering a large range of grammatical errors.
In addition, we also treat FlaCGEC as a diagnostic dataset for testing generalization skills and conduct a thorough evaluation of existing CGEC models. 
\end{abstract}

%%
%% The code below is generated by the tool at http://dl.acm.org/ccs.cfm.
%% Please copy and paste the code instead of the example below.
%%


%%
%% Keywords. The author(s) should pick words that accurately describe
%% the work being presented. Separate the keywords with commas.
%% A "teaser" image appears between the author and affiliation
%% information and the body of the document, and typically spans the
%% page.

%%
%% This command processes the author and affiliation and title
%% information and builds the first part of the formatted document.
\maketitle

\section{Introduction}
Writing grammatically correct Chinese sentences is difficult for learners studying Chinese as a Foreign Language (CFL) and even for native Chinese speakers due to its complex grammar rules.
Chinese Grammatical Error Correction (CGEC), aiming to detect and correct all grammatical errors in a sentence and produce an error-free sentence, has attracted intensive attention from researchers for its crucial value in many natural language processing scenarios such as writing assistant and search engine~\cite{duan:www2011,omelianchuk:etal2020,li:acl2021}.
Due to its profound significance, a surge of datasets have been observed~\cite{tseng:clp2015,zhao:nlpcc2018,rao:nlpt2018,zhang:naacl2022}.

However, most of these datasets~\cite{zhao:nlpcc2018,rao:nlpt2018} only provide correct sentences as the ground truth and have limitations in providing linguistic annotations to a CGEC method, which hinders the further improvement of a method.
Recent studies~\cite{zhang:naacl2022,ma:arxiv2022,xu:emnp20211} proposed CGEC datasets with grammatical error types.
However, their grammatical error types follow a shallow linguistic schema. 
A fine-grained linguistic schema widely covering the gramamtical points is demanded, which not only increases the interpretability of CGEC tasks, but also helps diagnose the CGEC methods~\cite{wang:tist2021,mita:acl2021}.

\begin{table}[t!]
    \centering
    \small
    \caption{Comparison of FlaCGEC with existing CGEC datasets.}
    \vspace{-0.3cm}
    \label{tab:datasets}
    \begin{tabular}{l c c c}
        \toprule  
        \textbf{Datasets} & \textbf{Annotation type} & \textbf{Type number} & \textbf{Source}  \\
        \midrule
        NLPCC~\cite{zhao:nlpcc2018} & Edits & 4 & CFL \\
        CGED~\cite{rao:nlpt2018, rao:CGED2020} & Edits & 4 & CFL \\
        CTC~\cite{zhao:ctc2022} & Edits & 3 & Native speaker\\
        MuCGEC~\cite{zhang:naacl2022} & Edits, Linguistic & $19$ & CFL  \\
        NaCGEC~\cite{ma:arxiv2022} & Edits, Linguistic & $26$ & Native speaker  \\
        FCGEC~\cite{xu:emnp20211} & Edits, Linguistic & $28$ & Native speaker \\
        \midrule
        FlaCGEC & Edits, Linguistic & $210$ & Native speaker \\
        \bottomrule
    \end{tabular}
    \vspace{-0.2cm}
    \vspace{-0.3cm}
\end{table}

To this end, we present \textbf{FlaCGEC}\footnote{Website: \url{https://github.com/hyDududu/FlaCGEC}}, a Chinese grammatical error correction dataset with fine-grained linguistic annotation.
We show an overall comparison of the differences between FlaCGEC and other datasets in Table~\ref{tab:datasets}.
% We then collect target sentences from the Chinese proficiency test to obtain a corpus with diverse grammatical points.
We first derive a linguistic schema from the grading standards textbook, where $78$ instantiated grammar points are organized in a deep hierarchical structure.
For data collection, we first collect target sentences from Chinese proficiency test for the sake of obtaining a corpus with diverse grammatical points.
Then we design edit rules to generate a set of enormous sentences for each target sentence and align the grammar points to corresponding edits.
Eventually, we obtain erroneous sentences covering diverse grammatical errors, which will be further verified by annotators.

We reproduce the state-of-the-art CGEC models and thoroughly evaluate them on FlaCGEC.
We discover there is a gap between the best baseline model and human performance, which indicates that FlaCGEC is still challenging.
Furthermore, we observe a significant performance drop of models that are trained on existing datasets. 
This reveals the distinction of FlaCGEC.
We also consider FlaCGEC as a diagnostic dataset for analyzing the existing CGEC models and discover that current models have poor generalization capability over diverse grammatical types and struggle to correct sentences with the complicated syntax as well as special usage of grammar points.
We hope FlaCGEC dataset could provide a comprehensive challenge to encourage more contributions to CGEC tasks.

\section{Dataset}

\subsection{Annotation}

FlaCGEC dataset enables quantitative and comprehensive evaluation of CGEC methods.
It not only provides the target sentences as the golden standards, but also annotates the error types explicitly.
Following the M2 format annotation of general GEC datasets ~\cite{bryant:bea2019,ng:cnll2013}, we annotate data with (1) the span of grammatical erroneous context (2) the error type and (3) the corresponding correction.

To ensure the annotated error types are canonical and recognized by the standard syllabus, we apply the official grammatical types in \textit{Chinese Proficiency Grading Standards for International Chinese Language Education}\footnote{The textbook could be found in {\tiny \url{http://www.moe.gov.cn/jyb_sjzl/ziliao/A19/202111/W020211118507389477190.pdf}.}} for annotating.
This could help examine the performance of a CGEC method on specific error types and give fine-grained feedback to CGEC approaches.
We show an example with M2 format annotation of FlaCGEC below.

\vspace{-0.2cm}
% \vspace{-0.2cm}
\begin{table}[h!]
    \centering
\begin{tabular}{l}
 $\mathtt{[S]}$ \text{\begin{CJK*}{UTF8}{gbsn}章鱼有发达的神经系统，\textcolor{red}{为}人亲善。\end{CJK*}} \\
Translation: Octopuses have powerful neural \\ systems, \textcolor{red}{regarding} human kindly. \\
 $\mathtt{[T]}$ \text{\begin{CJK*}{UTF8}{gbsn}章鱼有发达的神经系统，对人亲善。\end{CJK*}}\\
Translation: Octopuses have powerful neural \\ systems,  treating human kindly. \\
$\mathtt{[A]}$ $11$ $\quad$ $11$ $\mid\mid\mid$ \text{S-Preposition{\tiny [\begin{CJK*}{UTF8}{gbsn}词类介词\end{CJK*}]}} $\mid\mid\mid$ \begin{CJK*}{UTF8}{gbsn}对\end{CJK*}\\
% $\mathtt{[A]}$ $11$ $\quad$ $11$ $\mid\mid\mid$ \text{S-Preposition{\tiny [\begin{CJK*}{UTF8}{gbsn}词类介词\end{CJK*}]}\begin{CJK*}{UTF8}{gbsn}：对\end{CJK*}-2} $\mid\mid\mid$ \begin{CJK*}{UTF8}{gbsn}对\end{CJK*}\\
\end{tabular}
% \vspace{-0.2cm}
\vspace{-0.3cm}
\end{table}
where the lines preceded by $\mathtt{[S]}$ and $\mathtt{[T]}$ represent the source sentence and target sentence, respectively.
$\mathtt{[A]}$ goes ahead of an annotation, which consists of start token index, end token index, instantiated grammar point as well as edit, and correction.
The above annotation indicates that to correct the sentence, preposition ``\begin{CJK*}{UTF8}{gbsn}为\end{CJK*}'' should be changed to ``\begin{CJK*}{UTF8}{gbsn}对\end{CJK*}''.
% , which belongs to difficulty level 2.
We allow a single source sentence to contain multiple grammatical errors as presented in Table~\ref{tab:case}.
%as presented in Appendix~\ref{ap:more_examples}.\hycomment{remove or modify the statement?}

\subsection{Data Collection}

\begin{figure}[t!]
 \centering
 \includegraphics[width=0.33\textwidth]{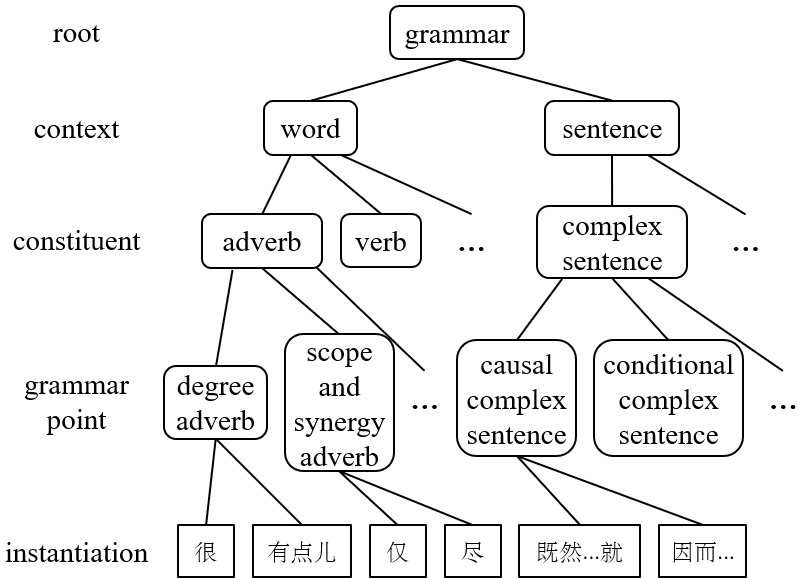}
 \centering
 \vspace{-0.3cm}
 \caption{Hierarchical structure of linguistic schema. }
 \vspace{-0.2cm}
 \label{fig:stucture}
 \vspace{-0.2cm}
 \vspace{-0.3cm}
\end{figure}

\noindent \textbf{Linguistic Schema.} 
% Our ultimate goal is to construct a CGEC dataset with fine-grained linguistic annotation.
Our goal is to construct a CGEC dataset with fine-grained linguistic annotation.
Unlike previous studies which manually designed linguistic schema for grammatical error types ~\cite{ma:arxiv2022}, we employ the schema defined in the grading standards textbook to specify the grammar points.
Figure~\ref{fig:stucture} demonstrates the deep hierarchical structure of our linguistic schema.
% For each grammar point, there are different words and phrases as instantiation.
Each grammar point has various words and phrases as examples.
% In the following section, we annotate grammatical errors based on the above schema.
In the next section, we annotate grammatical errors using the above schema.

\noindent \textbf{Target Sentence Collection.}
To collect target sentences, we utilize reading corpus from HSK exam~\cite{cui:book2013,zhang:book2011}, which is an official Chinese proficiency test. 
HSK corpus contains passages, where the sentences generally follow the above grammatical schema.
It is worth noting that previous CGED datasets~\cite{rao:nlpt2018} also employed HSK exam, but they utilize sentences wrtitten by CFL in HSK exam, which is more likely to reveal the limited grammatical errors from CFL.
By contrast, our study creates more complicated sentences based on the standard reading corpus featured with varying grammatical errors.

Therefore, we extract all the passages in HSK corpus via OCR and chunk them into sentences.
Then we either write regular expressions of close grammar points like \textit{pronouns} to annotate sentences or collect the illustrative examples with annotation from textbook.
These sentences are treated as the initial data pool.
Next, we iteratively train a tagging model to predict the grammar points contained in a sentence by adding the predicted sentences with high confidence into the data pool as augmented data~\cite{feng:acl2021}.
Eventually, we collect a set of target sentences from HSK corpus that are annotated with grammar points.

\noindent \textbf{Source Sentence Generation.}
Following the traditional methods~\cite{ma:arxiv2022} that automatically generate large-scale training data containing grammatical errors, we generate erroneous sentences via the following edits:
\begin{itemize}
\item \textbf{Removing words} means we randomly remove words of certain grammar points from the sentences such that some grammatical components are missing.
    We denote this edit type as ``M''.    
\item \textbf{Substituting words} means we randomly replace words of certain grammar types in the sentences with another word of the same grammar types such that the collocation of the sentence is not appropriate.
    We denote this edit type as ``S''.    
\item \textbf{Reordering words} means we randomly reorder the words of certain grammar points in the sentence leading to incorrect sentence syntax.
    We denote this edit type as ``W''.
\end{itemize}

Starting from the collected target sentences with annotated grammar points, we first do Chinese word segmentation via Jieba toolkit\footnote{\tiny \url{https://github.com/fxsjy/jieba}.}.
% For each word that is related to a grammar point, we randomly perform one or multiple of the above edits and there is still a chance that we keep the word unchanged.
For words related to grammar points, we randomly perform one or multiple of the above edits, with a possibility of leaving the word unchanged.
This results in multiple edits on a single target sentence and even multiple edits on the same words.
Eventually, we collect about $12,568$ source sentences. 
Each source sentence is associated with its target sentence as well as the corresponding M2 annotation.

\noindent \textbf{Bad Case Filtering.}
Random combinations of edits and grammar points result in a large number of candidate instances but some of them are improperly applied resulting in invalid source sentences.
Then we employ native speakers to filter out bad cases.
Bad cases are identified if: 1) too many errors exist in a single source sentence leading to confusing semantics which cannot be recovered even by native speakers; 2) the grammatical errors in sentences rarely exit in real-life scenarios and cannot be reproduced by annotators.

Specifically, we invite $13$ Chinese postgraduate students to filter the bad cases.
We write an annotation guideline to help the annotators better understand the annotation task.
Meanwhile, we provide intensive training to them before annotation.
Annotators should determine whether an example is a bad case based on the above judgment criterion and give a range of scores depending on the matching degree to the above criterias.

% To ensure the high-quality of our dataset, we select a senior annotator to review.
To ensure dataset quality, we select a senior annotator to review.
For each batch of annotated data, the senior annotator randomly samples instances to review.
% For each batch of data annotated by an annotator, the senior annotator randomly samples some instances to review. 
% If the disagreement degree of the annotations exceeds a threshold, the batch of data is re-assigned to a new annotator until it is accepted.
If the annotation disagreement exceeds a threshold, the batch is reassigned to a new annotator until agreement is reached.
After that, we filter out the bad cases, keep the rest and randomly split the data into training, development, and test sets with an $8:1:1$ ratio and without target sentences overlapping for data split.
% After that, we filter out the bad cases, keep the rest and randomly split the data into training, development, and test sets with a ratio of $8:1:1$ and without target sentences overlapping for data split.
This results in our FlaCGEC dataset.

\subsection{Data Analysis}

\begin{table}[t!]
    \centering
    \small
    \caption{Statistics and properties of FlaCGEC dataset.}
    \vspace{-0.3cm}
    \label{tab:statistics}
\begin{tabular}{l | l l l}
        \toprule
        \textbf{Properties} & \textbf{Train} & \textbf{Dev} & \textbf{Test} \\
        \midrule
        \#Sentences & 10,804 & 1,334 & 1,325 \\
        Average source sentence length & 35.09 & 34.76 & 35.83 \\
        Average target sentence length & 35.59 & 35.29 & 36.34 \\
        % \#Tokens per sentence & \\
        \#Edits per sentence & 1.72 & 1.69 & 1.71 \\
        \#Grammar points & 77 & 69 & 72 \\
        \bottomrule
    \end{tabular}
    % \vspace{-0.2cm}
    \vspace{-0.3cm}
\end{table}

\begin{table}[t!]
    \centering
    \small    
    \caption{Detection and correction results evaluated on FlaCGEC test set with different training data.
    $\Delta$ denotes the decrease percentage of $F_{0.5}$ of the current setting based on FlaCGEC $\rightarrow$ FlaCGEC.
    }
    \vspace{-0.3cm}
    \resizebox{0.999\linewidth}{!}{
\begin{tabular}{l | c | c c c c | c c c c}
        \toprule  
        \multirow{2}{*}{\textbf{Train Data} $\rightarrow$ \textbf{Test Data}} & \multirow{2}{*}{\textbf{Model}}  & \multicolumn{4}{c}{\textbf{Detection}} & \multicolumn{4}{c}{\textbf{Correction}} \\
         & & $R$ & $P$ & $F_{0.5}$ & $\Delta$ & $R$ & $P$ & $F_{0.5}$ & $\Delta$\\
        \toprule  
        % \bottomrule
        \multirow{3}{*}{FCGEC $\rightarrow$ FlaCGEC} & GECToR-Chinese & $9.95$ & $34.01$ & $22.92$ & $65.90$ & $5.91$ & $14.49$ & $11.23$ & $56.98$   \\
        %& TMTC \\
        & Chinese BART & $10.58$ & $19.11$ & $16.46$ & $48.08$ & $9.54$ & $12.49$ & $11.76$ & $31.62$\\
        & EBGEC & $3.31$ & $19.21$ & $9.80$ & $86.73$ & $3.17$ & $13.26$ & $8.11$ & $87.50$ \\
        \midrule
        \multirow{3}{*}{CTC $\rightarrow$ FlaCGEC} & GECToR-Chinese & $26.39$ & $51.19$ & $43.09$ & $38.59$ & $20.03$ & $29.77$ & $27.13$ & $32.46$\\
        % & TMTC \\
        & Chinese BART & $26.43$ & $34.52$ & $32.53$ & $26.32$ & $24.35$ & $23.98$ & $24.05$ & $12.67$\\
        & EBGEC & $3.72$ & $33.19$ & $12.85$ & $82.60$ & $3.68$ & $24.36$ & $11.48$ & $82.30$ \\
        \midrule
        \multirow{3}{*}{FlaCGEC $\rightarrow$ FlaCGEC} & GECToR-Chinese & $66.62$ & $72.95$ & $71.59$ & $-$ & $50.03$ & $47.75$ & $48.19$ & $-$ \\
        & Chinese BART & $52.50$ & $51.84$ & $51.97$ & $-$ & $43.48$ & $35.98$ & $37.27$ & $-$ \\
        & EBGEC & $\mathbf{79.56}$ & $\mathbf{72.55}$ & $\mathbf{73.85}$ & $-$ & $\mathbf{75.33}$ & $\mathbf{62.70}$ & $\mathbf{64.87}$ & $-$ \\    
        \midrule
        \multicolumn{2}{c|}{Human} & 78.64 & 86.93 & 85.14 & & 63.95 & 73.72 & 71.53 & \\
        \bottomrule
    \end{tabular}}
    \vspace{-0.2cm}
    \label{tab:cgec_results}
\end{table}

\begin{figure}[t!]
 \centering
 \includegraphics[width=0.33\textwidth]{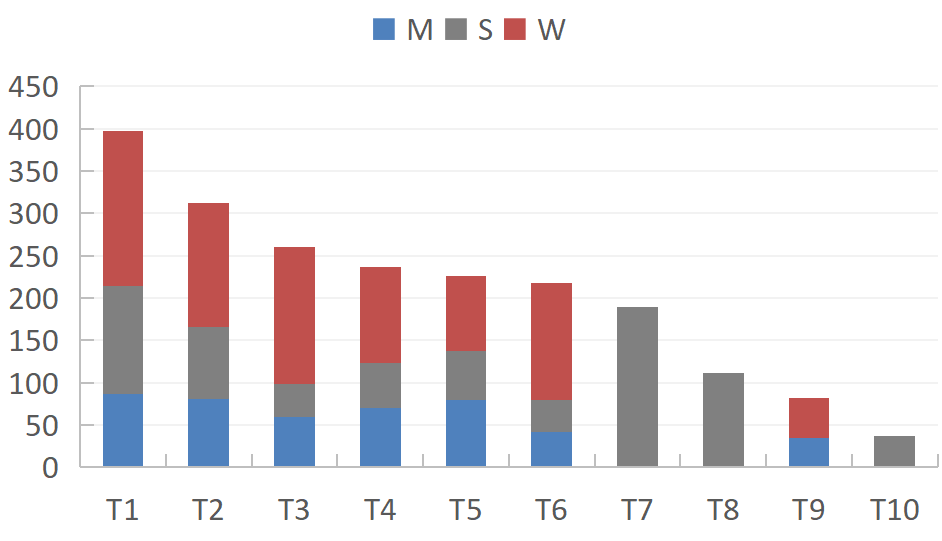}
 \centering
 \vspace{-0.3cm}
 \caption{Distribution of edit types on selected grammar points in FlaCGEC.
 T1: \textit{passive sentence}; T2: \textit{preposition for places}; T3: \textit{successive complex sentence}; T4: \textit{casual complex sentence}; T5: \textit{comparative sentence}; T6: ``is''-\textit{sentence}; T7: \textit{interrogative sentence}; T8: \textit{declarative sentence}; T9: \textit{preposition for time}; T10: \textit{exclamatory sentence}.}
 \vspace{-0.2cm}
 \label{fig:distribution}
 \vspace{-0.2cm}
 % \vspace{-0.3cm}
\end{figure}

We report the statistics of FlaCGEC dataset in Table~\ref{tab:statistics}.
We can see that FlaCGEC dataset provides a great number of erroneous sentences for training a good CGEC model.
The varying error types are included in the dataset and they are evenly distributed over training, development and test sets.
We display the distribution of edit types on selected grammatical types in FlaCGEC dataset in Figure~\ref{fig:distribution}.
We sample 10 grammar points from our linguistic schema and display the distribution of edits on these grammar points in FlaCGEC dataset.
We notice that there is a variance of the frequencies between grammar points (e.g., more errors exist in \textit{passive sentences} than \textit{exclamatory sentences}).
There is a correlation between edits and grammar points (e.g., \textit{exclamatory sentences} only have \textit{substitution} edits while \textit{prepositions for time} has a lack of \textit{substitution} edits).
This is because some combinations of grammar points and edits are invalid and recognized as bad cases.
This could help a GEC model learn the nature of language expression~\cite{xu:emnp20211}.

\section{Experiments}

\subsection{Experimental Setup}

To test the performance of cutting-edge CGEC approaches on our FlaCGEC dataset, we adopt three mainstream CGEC models: GECToR-Chinese~\cite{zhang:naacl2022}, Chinese BART~\cite{zhang:naacl2022}\footnote{\tiny \url{https://github.com/HillZhang1999/MuCGEC}} and EBGEC~\cite{kaneko:EBGEC2022}\footnote{\tiny \url{https://github.com/kanekomasahiro/eb-gec}.}.
We use the public source codes of these benchmark models, maintain their official hyperparameters, and perform experiments with various settings, which are illustrated in detail in the following sections.

In terms of evaluation metrices, we refer to the CLTC2022 shared task~\cite{CLTC2022}\footnote{\tiny \url{https://github.com/blcuicall/CCL2022-CLTC}} and employ \textbf{MaxMatch} (M$_2$) scorer~\cite{dahlmeier:bea@naacl2013} for evaluation.
We treat a detection prediction as correct if the predicted start token index and end token index are identical to the ground truth.
On top of that, if the edit is identical to the ground truth, the correction prediction is correct.
We report standard micro Precision, Recall, and F$_{0.5}$ score to evaluate the performance.

\begin{figure}[t!]
 \centering
 \includegraphics[width=0.33\textwidth]{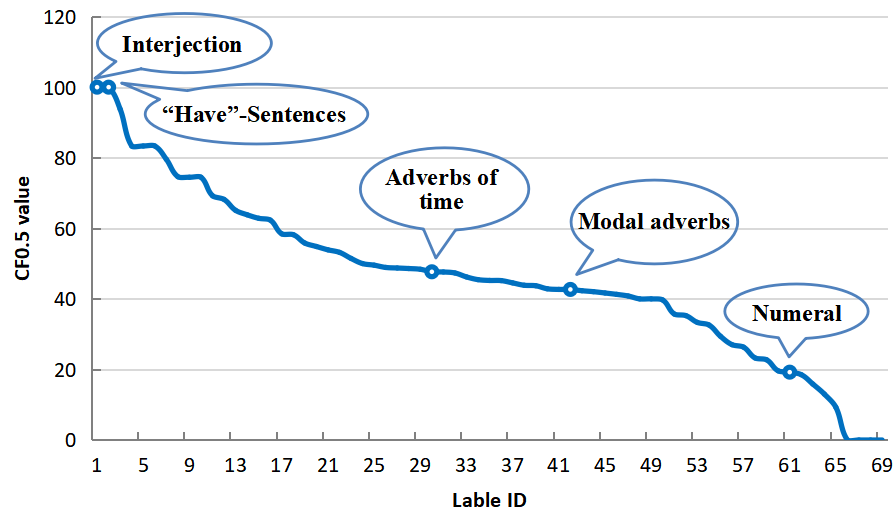}
 \centering
 \vspace{-0.3cm}
 \caption{Correction $F_{0.5}$ of GECToR-Chinese on FlaCGEC test set over fine-grained grammatical points.}
 \vspace{-0.2cm}
 \label{fig:fine_grained}
\end{figure}

\begin{table}[t!]
    \centering
    \small
    \caption{Some error cases predicted by GECToR-Chinese. $\mathtt{[P]}$ goes ahead of a predicted sentence.}
 \vspace{-0.3cm}
    \label{tab:case}
    \resizebox{0.99\linewidth}{!}{
\begin{tabular}{l }
        \toprule  
        Modal adverb{\tiny [\begin{CJK*}{UTF8}{gbsn}情态副词 \end{CJK*}]}: \begin{CJK*}{UTF8}{gbsn}就\end{CJK*} \\
        \midrule
        $\mathtt{[S]}$ \begin{CJK*}{UTF8}{gbsn}尊重和被尊重，\textcolor{red}{就}站在镜子前面，…\end{CJK*} \\
        Translation: Respect and being respected, just standing in front of a mirror, ...\\
        $\mathtt{[T]}$ \begin{CJK*}{UTF8}{gbsn}尊重和被尊重，就好像站在镜子前面，…\end{CJK*}\\
        Translation: Respect and being respected, just like standing in front of a mirror, ...\\
        $\mathtt{[P]}$ \begin{CJK*}{UTF8}{gbsn}尊重和被尊重，就\textcolor{red}{好}站在镜子前面，…\end{CJK*}\\
        Translation: Respect and being respected, just \textcolor{red}{have to} stand in front of a mirror, ...\\
        \midrule
        Conjunctions for connecting clauses\begin{CJK*}{UTF8}{gbsn}{\tiny [介词连接分句]}\end{CJK*}:\begin{CJK*}{UTF8}{gbsn}虽然...但是\end{CJK*} \\
        Negative adverb{\tiny [\begin{CJK*}{UTF8}{gbsn}否定副词\end{CJK*}]}:\begin{CJK*}{UTF8}{gbsn}没有\end{CJK*} \\
        \midrule
        $\mathtt{[S]}$ \begin{CJK*}{UTF8}{gbsn}\textcolor{red}{但有没}受到老板的责备，\textcolor{red}{而且}他心里很失落。\end{CJK*} \\
        Translation: \textcolor{red}{But did he} not receive the blame from his boss, and he is upset  \\
        $\mathtt{[T]}$ \begin{CJK*}{UTF8}{gbsn}虽然没有受到老板的责备，但是他心里很失落。\end{CJK*}\\
        Translation: Even though he did not receive the blame from his boss, he is upset.\\
        $\mathtt{[P]}$ \begin{CJK*}{UTF8}{gbsn}虽然\textcolor{red}{有}受到老板的责备，但是他心里很失落。\end{CJK*}\\
        Translation: Even though he \textcolor{red}{received} the blame from his boss, he is upset.\\
        \bottomrule
    \end{tabular}}
    \vspace{-0.2cm}
    % \vspace{-0.3cm}
\end{table}

\subsection{Difficulty of FlaCGEC}

To understand the difficulty of the FlaCGEC dataset, we conduct a set of experiments on FlaCGEC.
The results are presented in Table~\ref{tab:cgec_results} and we have the following observations:

(1) Regarding human evaluation, we hire $4$ native speakers to annotate the test set.
As shown in Table~\ref{tab:cgec_results}, FlaCGEC dataset is also challenging for humans due to the wide range of grammatical errors it examines.
Overall, the human evaluation yields high precision but relatively low recall.
This observation is similar to prior study~\cite{xu:emnp20211}.

(2) Amongst the models trained on FlaCGEC dataset, EBGEC obtains the best result for detection and correction.
We speculate that this is because EBGEC is able to correct the grammatical errors by referring to the most similar training example while the other two methods entirely rely on the contexts.
However, EBGEC cannot play full of the advantage when it encounters a gap between training and test data.
In comparison, GECToR-Chinese shows more robust results in the three experimental settings, such that we conduct a set of analysis based on GECToR-Chinese in the next section.
Overall, there is still a significant gap of performance between the best model with humans on FlaCGEC.

(3) When we train the benchmark models on the other two datasets, we notice that their performance all drops a lot.
This reflects that FlaCGEC dataset contains a great number of grammatical errors that are not involved in other datasets. 
Comparing FCGEC and CTC datasets, FlaCGEC is more likely to have similar grammatical errors as those in CTC dataset.
We investigate and notice that FCGEC examines more on the correct syntactic construction of sentences while our dataset focuses more on the accurate discrimination of grammatical usage in sentences.

\subsection{More Analysis}

\noindent \textbf{Analysis of Fine-grained Performance.} We display the experimental results of GECToR-Chinese over fine-grained grammar points in Figure~\ref{fig:fine_grained}.
We obverse there is a variance of correction $F_{0.5}$ over fine-grained grammatical points.
The best results are close to $1$ while the worst results approach $0$.
Specifically, certain grammatical errors like \textit{interjection}, \textit{``have''-sentences} can be easily solved by GECToR-Chinese.
But it has difficulty in correcting errors related to \textit{numeral}.
This may be because \textit{numeral} has more flexible usage in Chinese expressions, which requires a deep understanding of the sentences and strong generalization capability of CGEC models.

\noindent \textbf{Case Study.}
We show some error cases in Table~\ref{tab:case}. 
Case 1 has grammar errors related to \textit{modal adverb}, which is not easy to be detected as a CGEC model needs to understand the context and know it is a metaphor.
GECToR-Chinese succeeds to understand the sentence and accurately detects the spans to be corrected.
But it fails to correct it by inserting a wrong modal adverb.
% For Case 2, a CGEC model needs to first understand the semantics of this sentence then detect the misuse of the conjunction as well as the contexts required for the conjunction.
In Case 2, a CGEC model must grasp sentence semantics, detect conjunction misuse, and understand the necessary contextual information.
GECToR-Chinese accurately corrects the conjunction but unsuccessfully comprehends the statement and the predicted sentence presents an invalid progressive description.
This indicates that CGEC models suffer more when special usage of grammar is examined.

\begin{table}[t!]
    \centering
    \small
    \caption{Detection and correction $F_{0.5}$ of PLMs evaluated on sampled FlaCGEC test data under zero-shot setting.}
    \vspace{-0.3cm}
    \label{tab:ChatGPT_results}
    % \resizebox{0.45\textwidth}{!}{
    \begin{tabular}{l | c c c | c c c }
        \toprule  
        \textbf{PLMs} & \multicolumn{3}{c}{\textbf{Detection}} & \multicolumn{3}{c}{\textbf{Correction}} \\
         & $R$ & $P$ & $F_{0.5}$ & $R$ & $P$ & $F_{0.5}$ \\
        \toprule  
        % \bottomrule
        ChatGPT & 39.36 & 23.12 & 25.30 & 25.18 & 11.36 & 12.76 \\
        GPT-3 & 39.89 & 33.19 & 34.34 & 23.02 & 15.31 & 16.41 \\
        \bottomrule
    \end{tabular}
    % }
    \vspace{-0.2cm}
    \vspace{-0.2cm}
\end{table}

\noindent \textbf{Zero-shot Transfer on FlaCGEC.}
Recently, Large-scale Language Models (LLMs)~\cite{devlin:naacl2019,brown:nips2020} have shown to be effective in \textit{few-shot} or \textit{zero-shot} scenarios.
To understand whether LLMs have the capability to solve FlaCGEC dataset under zero-shot setting, we employ ChatGPT\footnote{{\tiny \url{https://openai.com/blog/chatgpt/}}} and GPT-3 with the following prompt: 

\noindent Prompt($\mathbf{x}$) = \textit{I will show you a Chinese sentence with grammatical errors, please show me the correct sentence.
The wrong sentence is $\mathbf{x}$}.

We evaluate the generated sentences with M$_2$ and present the results in Table~\ref{tab:ChatGPT_results}.
From the results, we observe that the simple prompt could lead to remarkable gains in CGEC tasks.
But the results are still significantly worse than the state-of-the-art on both detection and correction, which are $73.85\%$ and $64.87\%$, respectively.
This is because LLMs recover basic semantics of sentences but neglect the accurate discrimination of the grammar points.

\section{Conclusions}

In this paper, we introduce FlaCGEC, a CGEC dataset with fine-grained linguistic annotation.
We conduct a thorough evaluation of cutting-edge CGEC methods showing that our dataset is challenging and could provide environments to test diverse generalization abilities of CGEC methods.
With FlaCGEC, current CGEC methods can be trained and diagnosed to improve performance continuously.

%%
%% The acknowledgments section is defined using the "acks" environment
%% (and NOT an unnumbered section). This ensures the proper
%% identification of the section in the article metadata, and the
%% consistent spelling of the heading.

\begin{acks}
The authors would like to thank the anonymous reviewers for
their insightful comments. 
This work was supported by East China Normal University (2022ECNU—WHCCYJ-31), Natural Science Foundation of China (Project No. 61977026) and Shanghai Pujiang Talent Program (Project No. 22PJ1403000).
\end{acks}

%%
%% The next two lines define the bibliography style to be used, and
%% the bibliography file.
\bibliographystyle{ACM-Reference-Format}
\balance
\bibliography{sigconf_main}

\end{document}